\documentclass[10pt,twocolumn,letterpaper]{article}

\usepackage{cvpr}      

\usepackage{graphicx}
\usepackage{amsmath}
\usepackage{amssymb}
\usepackage{booktabs}
\usepackage{subcaption}
\usepackage{caption}
\usepackage{xcolor}
\usepackage {appendix}
\definecolor{demphcolor}{RGB}{144,144,144}
\newcommand{\demph}[1]{\textcolor{demphcolor}{#1}}

\usepackage[pagebackref,breaklinks,colorlinks]{hyperref}

\newcommand{\modelname}{DDCap}

\usepackage[capitalize]{cleveref}
\crefname{section}{Sec.}{Secs.}
\Crefname{section}{Section}{Sections}
\Crefname{table}{Table}{Tables}
\crefname{table}{Tab.}{Tabs.}

\def\cvprPaperID{12477} 
\def\confName{CVPR}
\def\confYear{2023}

\begin{document}
\title{Exploring Discrete Diffusion Models for Image Captioning}

\author{Zixin Zhu$^1$\thanks{Equal. $^\dag$Corresponding authors. Zixin and Yixuan are interns at MSRA.}
\quad Yixuan Wei$^2$\textsuperscript{*} \quad Jianfeng Wang$^3$ \quad Zhe Gan$^3$ \quad Zheng Zhang$^3$
\quad Le Wang$^{1,\dag}$ \quad Gang Hua$^1$  \\
\quad Lijuan Wang$^3$ \quad Zicheng Liu$^3$
\quad Han Hu$^{3,\dag}$ \\
{$^1$Xi'an Jiaotong University} \quad  {$^2$Tsinghua University} \\
{$^3$Microsoft}  \\
\small{\texttt{zhuzixin@stu.xjtu.edu.cn}} \\
\small{\texttt{lewang@xjtu.edu.cn}} \\
\small{\texttt{\{t-yixuanwei,jianfw,zhe.gan,zhez,lijuanw,zliu,hanhu\}@microsoft.com}}
}
\maketitle

\begin{abstract}
The image captioning task is typically realized by an auto-regressive method that decodes the text tokens one by one.
We present a diffusion-based captioning model, dubbed the name \modelname\xspace, to allow more decoding flexibility.
Unlike image generation, where the output is continuous and redundant with a fixed length, texts in image captions are categorical and short with varied lengths. 
Therefore, naively applying the discrete diffusion model to text decoding does not work well, as shown in our experiments.
%
To address the performance gap, we propose several key techniques including best-first inference, concentrated attention mask, text length prediction, and image-free training. 
On COCO without additional caption pre-training, it achieves a CIDEr score of 117.8, which is +5.0 higher than 
the auto-regressive baseline with the same architecture in the controlled setting. 
It also performs +26.8 higher CIDEr score than the auto-regressive baseline  (230.3 v.s. 203.5) on a caption infilling task.
With 4M vision-language pre-training images and the base-sized model, we reach a CIDEr score of 125.1 on COCO, which is competitive to the best well-developed auto-regressive frameworks. The code is available at \url{https://github.com/buxiangzhiren/DDCap}.

\end{abstract}
\section{Introduction}
\label{sec:intro}
Diffusion models~\cite{ho2020denoising,song2020denoising} have been successfully exploited for image generation~\cite{dhariwal2021diffusion,nichol2021glide,ramesh2022hierarchical,saharia2022photorealistic,rombach2022high}, producing results with high fidelity. However, there is limited work, if at all, exploring diffusion models for text generation such as that in image captioning~\cite{vinyals2015show,karpathy2015deep}, where natural language description is generated to describe an image. In view of this under-exploration, in this work, we aim to conduct a systematic study on how we may adopt diffusion models for accurate image captioning. 

Our first attempt of naively applying either continuous or discrete diffusion models for image captioning did not work well, producing results that are much more inferior to state-of-the-art results produced from mainstream auto-regressive models pre-trained on millions of data and even billions of data, {\em i.e.}, OSCAR~\cite{li2020oscar}, VinVL~\cite{zhang2021vinvl}, UFO~\cite{wang2021ufo}, ViTCap~\cite{fang2022injecting}, SimVLM~\cite{wang2021simvlm}, and GIT~\cite{wang2022git}, to name a few. This motivated us to conduct a careful gap analysis. 
\begin{figure}[t]
   \centering
   \includegraphics[width=1\linewidth]{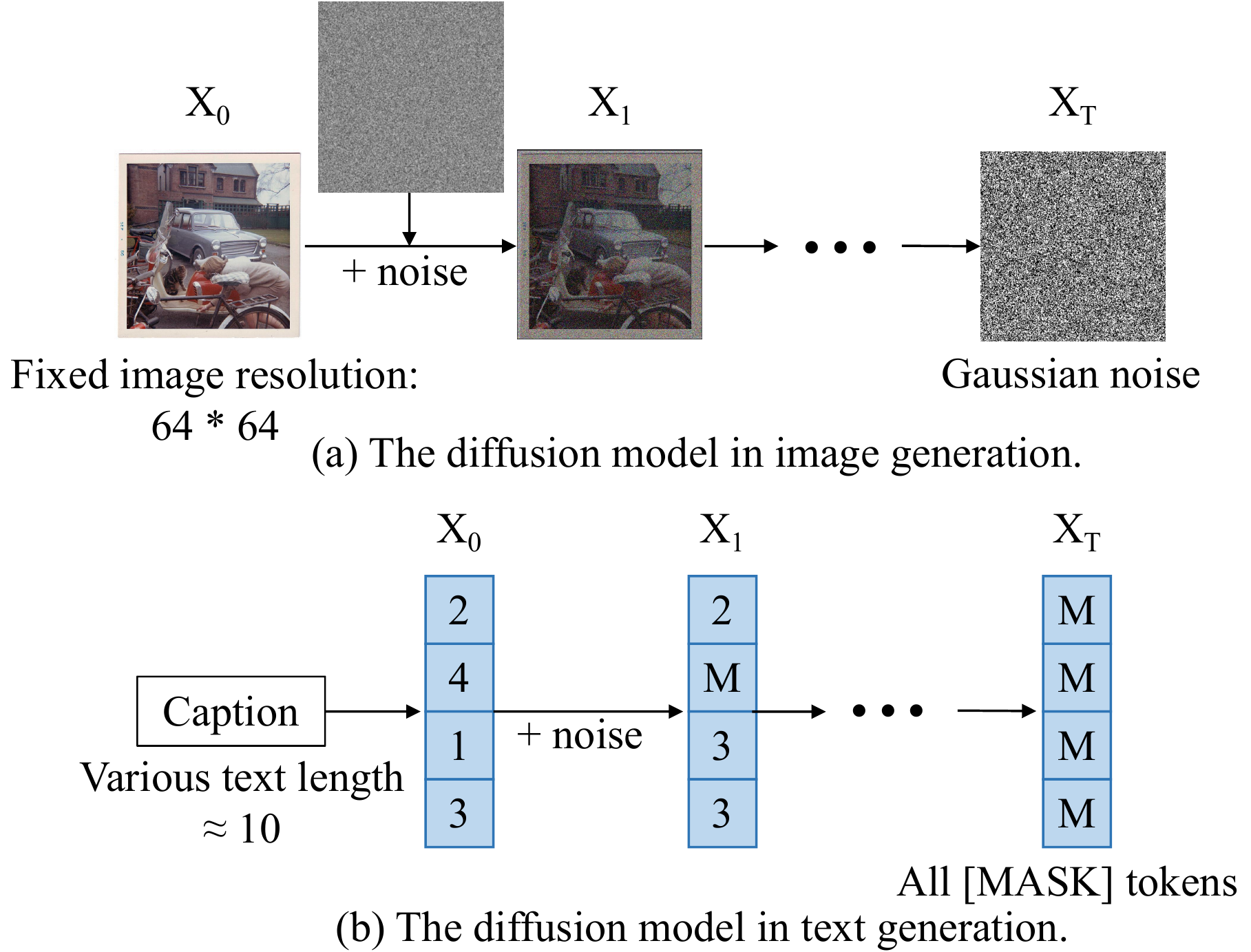}

   \caption{(a) Image generation: the noise is continuous and redundant with a fixed length (b) Text generation: the noise is categorical and short with varied lengths.}
   \label{fig:imgtext}
\end{figure}

As shown in Figure \ref{fig:imgtext}, the diffusion model exhibits different natures
between the image generation and the text generation.
On one hand, in image generation, the output images are {\em continuous and redundant} with a {\em fixed length}, and the information residing in the output is {\em redundant}.
On the other hand, in image captioning, the output texts are {\em discrete and concise} with a {\em variable length}. 
The diffused image resolution is typically $64\times64=4096$, while the length of the image caption tokens can be around $10$.
Besides, in the inference procedure of a diffusion model, some correct words generated at the current step may flip to a wrong words because of furtherly added noises.

Based on the above analysis and in observation that text tokens are discrete in nature, we focused our research exploration on making discrete diffusion model produce accurate image captions. Our proposed model, namely DDCap, is specifically designed to address the above gaps. 
First of all, we add a network branch to predict the total token length to flexibly accommodate variable lengths of the texts. 
Secondly, since texts are discrete and concise in nature, we design a {\em concentrated attention mask} module so that we may adaptively concentrate on more informative tokens.

Thirdly,
we propose a first best inference strategy, where 
the top-$K_t$ recovered tokens from each diffusion step unchanged in subsequent diffusion steps. In other words, in each diffusion step, we only add noises to those tokens that are not marked as fixed in previous diffusion steps. The top-$K_t$ in the current diffusion step are selected only from these remaining tokens which had been added noises, and they will be fixed in subsequent steps. The number $K_t$ is roughly set as the total number of tokens we need to recover divided by the number of diffusion steps.        

Last but not least, we further design an image-free training technique,
motivated by the classifier-free diffusion model~\cite{tang2022improved}. In some training examples, we remove the image conditions and force the network to learn the text prior knowledge. 
This enables a trade-off to balance the dataset prior and the image condition. In inference, the text prior knowledge and image condition are appropriately combined to obtain better captions than just depend on images.
With these orchestrations, we pre-train the DDCap model with a CLIP image encoder on datasets with 4M images (roughly 10M image-text pairs in total). We are able to achieve a CIDEr score of 125.1 on the COCO dataset, which is competitive when compared with state-of-the-art auto-regressive model. In further exploration of the capacity of our proposed DDCap model, we hypothesize that our DDCap model considers more holistic contexts to generate the texts than the auto-regressive baseline, when generating the text captions.


We further verified our hypothesis by conducting experiments in an infilling tasks, which is illustrated in Figure~\ref{fig:visualfill}. 
In this task, we remove all adjectives in the ground truth, and require the model to predict the removed adjectives. 
As for evaluation metrics, we add the clip scores~\cite{hessel2021clipscore} for semantic matching. Our proposed DDCap model outperforms the auto-regressive baseline by a significant margin.

The contributions of this paper are hence summarized as follows.
\begin{itemize}
    \item We are the first to apply discrete diffusion models for image captioning, and provide the first evidence that diffusion models can achieve competitive performance to the best well-developed auto-regressive models.
    \item We propose four key designs: ($i$) length prediction is used to deal with the varied length issue in text generation; ($ii$) concentrated attention mask is used to extract compact text information without the interference of undesired noise; 
    ($iii$) best-first inference is proposed to reduce the chance of contaminating correctly generated tokens; and ($iiii$) image-free training to balance the information in the text prior and the image condition. 
    \item We further show the advantages of discrete diffusion models in a caption infilling task.
\end{itemize}

The remainder of the paper is organized as follows. We summarize related work in Section~\ref{sec:relwork}. The details of the proposed DDCap model is proposed in Section~\ref{sec:methods}. Extensive experimental results are reported in Section~\ref{sec:exp}, with both quantitative and qualitative discussions. We finally conclude in Section~\ref{sec:conc}.
\section{Related Work~\label{sec:relwork}}
\begin{figure*}[!t]
   \centering
   \includegraphics[width=1\linewidth]{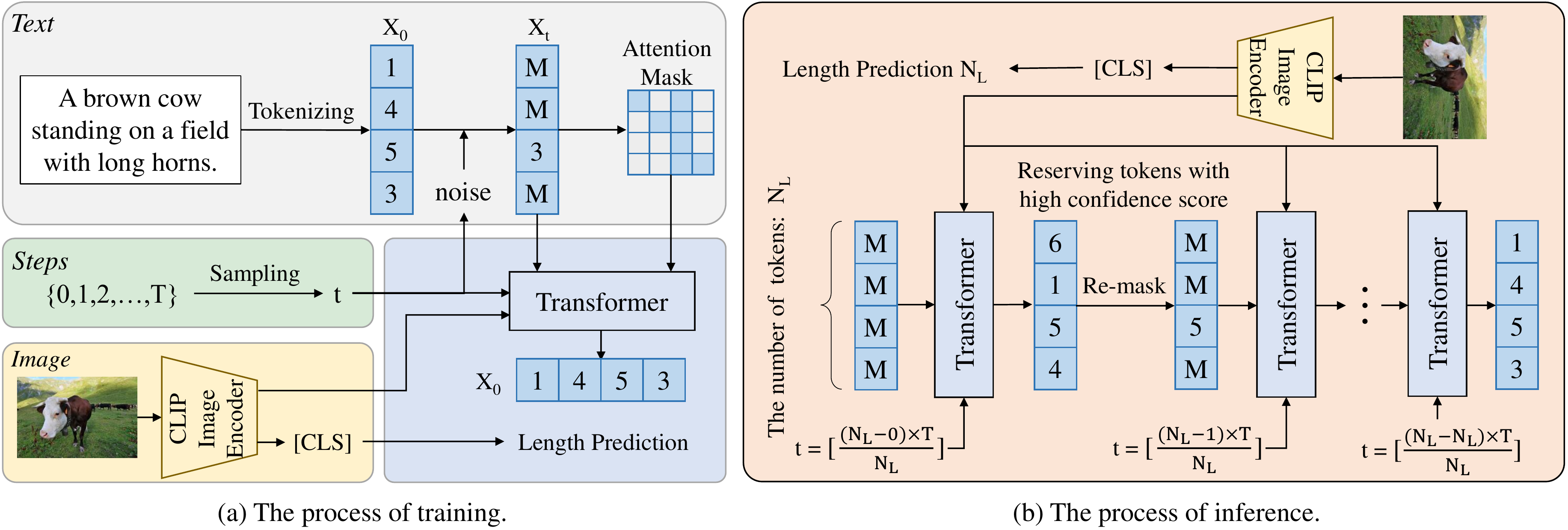}
   \caption{(a) A caption is first tokenized. Then, these tokens are added noise to become other text tokens or \texttt{[MASK]} tokens. The strength of noise depends on the sampled step $t$. Lastly, with the help of image tokens, we take these text tokens carried noise as the input of transformer to predict the pure text tokens. Meanwhile, we use a simple MLP to predict the length of the caption based on the \texttt{[CLS]} token. (b) According to the predicted length which denotes the number of \texttt{[MASK]} tokens, we take all \texttt{[MASK]} tokens as the input. Moreover, each step $t$ adjusted by the ratio of the predicted length and total noise step $T$ is sent to the Adaptive LayerNorm layer of the transformer. Image tokens are used to conduct cross-attention with text tokens. Combining three inputs, we can obtain the results of each step. The tokens with high confidence score will be retained instead of re-masking. }
   \label{fig:traininfer}
\end{figure*}
\paragraph{Image Captioning.}
Image captioning has been a longstanding task and recent years have witness great progress, especially with large-scale vision-language pretraining.
Early models~\cite{vinyals2015show,karpathy2015deep,rennie2017self,anderson2018bottom} use a visual encoder to extract visual features and apply recurrent neural network as decoder for caption generation. Later on, attention-based methods have been proposed to capture multimodal alignment~\cite{xu2015show,lu2017knowing,huang2019attention} and perform object relational reasoning~\cite{yao2019hierarchy,yao2018exploring}. Besides, researchers have explored to leverage semantic attributes~\cite{wu2016value,you2016image,gan2017semantic,yao2017boosting} and scene graphs~\cite{yang2019auto} for captioning. Enhanced attention-based models are further proposed to improve the performance, such as ORT~\cite{herdade2019image}, AoANet~\cite{huang2019attention}, M$^2$ Transformer~\cite{cornia2020meshed}, X-LAN~\cite{pan2020x}, and RSTNet~\cite{zhang2021rstnet}.
More recent efforts on image captioning are reflected from the perspective of novel model architecture design~\cite{fang2022injecting,pan2020x,nguyen2022grit,fei2022deecap} and the use of prior knowledge~\cite{meng2022object,hirota2022quantifying,wu2022difnet,li2022comprehending,kuo2022beyond}. For example, Fang \textit{et al.}~\cite{fang2022injecting} present a detector-free image captioning model with fully transformer architecture, 
and introduce a novel module to predict semantic concepts for high performance. 
Note that most previous works adopt a standard auto-regressive method to decode tokens one by one. 

For non-regressive approaches, masked language modeling is leveraged in~\cite{gao2019masked}
and reinforcement learning is applied in~\cite{guo2020non}. 
Instead, we focus on exploring diffusion models for image captioning, and achieve better performance.


\paragraph{Vision-Language Pre-training.} Recent advances in image captioning can be largely attributed to vision-language pre-training (VLP). Early VLP models~\cite{zhou2020unified,li2019visualbert,su2019vl,gan2020large} adopt a two-stage approach, where a pre-trained object detector is first used to extract regional features offline, and then a transformer is stacked on top for multimodal fusion. Prominent examples along this line include ViLBERT~\cite{lu2019vilbert}, LXMERT~\cite{tan-bansal-2019-lxmert}, UNITER~\cite{chen2020uniter}, OSCAR~\cite{li2020oscar}, VinVL~\cite{zhang2021vinvl}, and LEMON~\cite{hu2021scaling}. More recently, end-to-end VLP models that use convolutional networks and vision transformers as the image encoder are becoming the mainstream~\cite{kim2021vilt,li2021align,dou2021empirical,wang2022ofa,li2022blip,dou2022coarse,chen2022pali}. For example, SimVLM~\cite{wang2021simvlm} is pre-trained with a single prefix language modeling objective on large-scale image-text data, GIT~\cite{wang2022git}  simplifies the architecture as one image encoder and one text decoder under a single language modeling task, and CoCa~\cite{yu2022coca} is pre-trained with a mix of contrastive and generative losses. In this work, we also perform VLP for image captioning, but distinct from all the above works, we are the first to pre-train a discrete diffusion model for image captioning (and more generally, for text generation).


\paragraph{Diffusion Models.}
Diffusion models can be roughly divided into two categories: ($i$) continuous diffusion, and ($ii$) discrete diffusion.
Continuous diffusion models~\cite{ho2020denoising, song2020denoising, dhariwal2021diffusion,zhang2022gddim} add noise on the continuous-valued input or latent features. Recently, this family of models have achieved great successes in text-to-image generation, such as GLIDE~\cite{nichol2021glide}, DALLE-2~\cite{ramesh2022hierarchical}, Imagen~\cite{saharia2022photorealistic}, and Stable Diffusion~\cite{rombach2022high}. 
On the other hand, discrete diffusion~\cite{sohl2015deep}, especially its application to text generation, has been rarely studied, which is the focus of this work. Specifically, ImageBART~\cite{esser2021imagebart} and VQ-Diffusion~\cite{gu2022vector} propose to model the latent discrete code space of a VQ-VAE~\cite{razavi2019generating} by learning a parametric model using a conditional variant of DDPM~\cite{ho2020denoising}, for the task of text-to-image generation. 
For text generation, discrete diffusion has only shown some initial success on relatively toy problems~\cite{hoogeboom2021argmax, austin2021structured}. Recent work~\cite{li2022diffusion} has also tried to use continuous diffusion for text generation, but more for fine-grained control tasks. We are the first to apply discrete diffusion models for image captioning, and provide the first evidence that diffusion models can achieve competitive performance to the best well-developed auto-regressive models. 

\section{Methods}\label{sec:methods}


We present \modelname, which is a discrete diffusion model dedicated for image captioning task. This allows more decoding
flexibility, compared to the current dominant auto-regressive methods. For example, multiple tokens can be predicted simultaneously rather than one by one. The tokens from the left side can also depend on those on the right side, instead of a fixed left-to-right generation process. In Section~\ref{sec:ddcap_framework}, we first introduce the key idea of applying discrete diffusion models for the captioning task, and then in Section~\ref{sec:key_design}, we detail multiple techniques that play a crucial role to boost the performance. 

\subsection{Discrete diffusion model for image captioning}\label{sec:ddcap_framework}

Unlike continuous diffusion models which add token-level continuous noises individually to the RGB values or features of each token, discrete diffusion models perform a noising process at the sentence level, where mask tokens randomly replace regular tokens with noise levels represented by masking ratio.

\paragraph{Noising process}

We denote each text token in a caption as a discrete state, $x$. In the experiments, we use the BPE tokenizer~\cite{sennrich2015neural} in GPT-2 to generate the token identities of the sentence.

The noising process can be thought of as the Markov process, which gradually adds noises step-by-step. We denote the corresponding noisy token at Step $t$ as $x_t$. At each step, each token has a probability of $\gamma_t$ transiting to a special absorbing state of \texttt{[MASK]}.  If $x_{t-1}$ is not \texttt{[MASK]}, the transition probability vector from Step $t-1$ to $t$ is defined as
\begin{align}
    p(x_t | x_{t - 1}) = 
    \begin{cases}
    \alpha_t,\,\, x_t = x_{t - 1}; \\
    \gamma_t,\,\, x_t = \texttt{[MASK]}; \\
    1 - \alpha_t - \gamma_t, \,\,\text{otherwise}.
    \end{cases}
\end{align}
In other words, the each token of $\boldsymbol{x}_{t-1}$ has a probability of $\alpha_{t}$ to be unchanged and has a probability of $\gamma_{t}$ to be replaced by the $[\text{MASK}]$ token, leaving the probability of $\beta_{t} = 1 - \alpha_{t} - \gamma_{t} $ to be replaced by other tokens except the $[\text{MASK}]$ token in the vocabulary. If $x_{t-1}$ is a \texttt{[MASK]} token, the transition probability vector from Step $t-1$ to $t$ is defined
\begin{align}
    p(x_t | x_{t - 1} = \texttt{[MASK]}) = 
    \begin{cases}
    1,\,\, x_t =\texttt{[MASK]}; \\
    0,\,\, \text{otherwise}.
    \end{cases}
\end{align}

By these noising process, with a sufficient number of steps $T$, all tokens will become \texttt{[MASK]}.

\paragraph{Denoising process}
The denoising diffusion model starts with an all \texttt{[MASK]} sequence and gradually recovers the original signal step-by-step. A Transformer network is used to predict the reserve projection, i.e., $p_{\theta}(x_{t-1} | x_t, y)$, where $y$ is the image feature extracted from an image encoder, i.e., initialized from CLIP~\cite{radford2021learning} in our experiments. Cross-attention layers are employed to incorporate image features as conditions for the caption generation process. The time step $t$ is encoded as a sinusoidal positional embedding to guide the adaptive layer norm~\cite{gu2022vector} in the Transformer layers:
\begin{align}
    p &= t * \text{step}_{\text{scale}} / T,\\
    \text{PE}_{i} &= \begin{cases}
    sin(p/10000^{2i/d_{\text{model}}}), i < d_{\text{model} }/ 2 \\
    cos(p/10000^{2i/d_{\text{model}}}), i \ge d_{\text{model} } / 2,
    \end{cases}
    \label{eqn:sin}
\end{align}
where $\text{step}_{\text{scale}}$ is the wavelength, i.e., 8000 in our experiments, and $d_{\text{model}}$ is the hidden dimension.

\paragraph{Training} Unlike continuous diffusion models, where denoising networks typically predict noises, discrete diffusion models perform the noising process at the sentence-level through a masking strategy, and are difficult to directly predict noises. Inspired by~\cite{austin2021structured}, we directly predict the original text tokens $x_0$ instead:
\begin{equation}
\mathcal{L}_{x_0}=-\log p_{\theta}(x_0 | x_t, y).
\end{equation}

The training process is illustrated as the left part of Figure~\ref{fig:traininfer}. 

\paragraph{Inference} Starting from $x_{T}$, the model first predicts $x_{T-1}$, then predicts $x_{T-2}$, and so on. After $T$ steps, the discrete diffusion model produces the final result of $x_0$. 

The process of predicting $x_{t-1}$ from $x_{t}$ is computed as follows. First, $\hat{x}_{0}$ is estimated from $p_{\theta}(x_0 | x_t, y)$ via the trained denoising network. 
Then, $\{t-1\}$-step noise is added on the predicted $\hat{x}_{0}$ through a Markov chain to get $x_{t-1}$.

\subsection{Key Designs for Performance Improvement}\label{sec:key_design}
Empirically, with naive implementations, we observe significant worse performance with the diffusion model than the auto-regressive counterpart. We hypothesize that this is due to that text generation is very different from image generation. For example, the generated text lengths are different, while the generated image resolutions are normally fixed.
The information in the text is compact, while the image contains redundant information. 
The mask token provides almost no information, while the image is continuously noised into different levels. To address the above issues, we propose the following four techniques: ($i$) length prediction, ($ii$) concentrated attention mask, ($iii$) best-first inference and ($iiii$) image-free training.

\paragraph{Length Prediction.}


In auto-regressive methods, the text tokens are predicted from left to right one by one until a special \texttt{[EOS]} token. However, diffusion model has no concept of left-to-right or right-to-left directions. 
The performance is also much worse if we respect the \texttt{[EOS]}
token and treat only the left-side tokens as valid.
Instead of relying on the special token, we propose to add a network branch to specifically predict
the total token length. 
Specifically, we regard the \texttt{[CLS]} token from the CLIP image encoder as the feature for length prediction, implemented via a simple Multilayer Perceptron (MLP). Cross-entropy loss is used 
for training. Empirically, we also find it is beneficial to cut the gradient back to the 
image encoder.
Without the length as a prior, we may have to use a maximum token length to generate the captions.
Thus, another benefit is the speed-up with an appropriately predicted token length.

\paragraph{Concentrated Attention Mask (CAM).}

In our diffusion model, the absorbing \texttt{[MASK]} token carries almost no information,
and it can take a majority in the sequence. For instance, the denoising process starts from a full \texttt{[MASK]} sequence. With the bidirectional attention, the \texttt{[MASK]} tokens may overwhelm the attention layers. To make the tokens only concentrate on informative tokens, we propose the following changes on the attention mask: ($i$) normal text tokens do not depend on the \texttt{[MASK]} tokens; ($ii$) the \texttt{[MASK]} token does not depend on the other \texttt{[MASK]} tokens. 
Empirically, we find this novel attention mask design significantly boosts the performance.

\begin{table*}[!t]
  \small
  \centering
  \begin{tabular}{@{}c cccc|c|c|c|c|c@{}}
    \toprule   
    \#Row  & Best-first inference & CAM & Length Prediction & Image-free training & C & B@4 & M & R & S \\    
    \midrule
    a  &&&&& 20.6 & 7.4  & 18.8 & 34.6 &12.3  \\ 
    b  &&$\checkmark$&&& 43.6 & 11.4  & 20.5 & 39.1 & 14.4  \\ 
    c   &$\checkmark$ &&&& 45.2 & 20.3  & 26.9 & 47.3 & 21.3  \\
    d &$\checkmark$&&$\checkmark$& & 92.6 & 27.3  & 25.4 & 51.8 & 18.7\\
    e  &$\checkmark$&$\checkmark$&&& 97.5 & 28.2  & 28.1 & 54.0 & \textbf{21.7} \\
    f  &$\checkmark$&$\checkmark$&$\checkmark$&& 116.7 & 34.6  & 28.1 & 57.4 & 21.5\\
    g &$\checkmark$&$\checkmark$&$\checkmark$&$\checkmark$& \textbf{117.8} & \textbf{35.0}  & \textbf{28.2} & \textbf{57.4} & \textbf{21.7}\\
    \bottomrule
  \end{tabular}
  \caption{Ablation study on the effectiveness of each component described 
  in Sec.~\ref{sec:key_design}. CAM: concentrated attention mask.
  }
  \label{tab:component}
\end{table*}


\paragraph{Best-first Inference.}
During inference, the text token can be recovered earlier than the last step. 
To reduce the chance of contaminating these correct tokens, we propose to
keep the top-$K^t$ recovered tokens unchanged at each step. 
Let $N_L$ be the predicted token length. 
If $N_L \le T$ ($T$ is the total diffusion steps during training), we also reduce
the inference diffusion steps to $N_L$, such that at each step, one extra token is recovered. 
Meanwhile, the step index $t$, which is used in the adaptive layer norm, is also shrunk proportionally.
If the predicted token length is larger ($N_L > T$), at each step $t$ (from $T$ to $1$), we keep 
$K^t = \lfloor N_L (T-t + 1)/T  \rfloor - \lfloor N_L (T-t)/T  \rfloor$ tokens unchanged.

\paragraph{Image-free Training.}
Generating the text is the key of the image captioning task. To focus more on the text modeling part, 
we propose an image-free strategy to enhance the weight of text modeling, which is motivated from the classifier-free diffusion~\cite{tang2022improved}. 
Specifically,
with a probability of $r$ (0.2 in experiments),
the image features are replaced with
a trainable embedding $f$, and
the diffusion model calculates $p_\theta(x_0|x_t, f)$ without image conditions. The loss then becomes
\begin{align}
    \mathcal{L}_{x_0}' = -\log p_{\theta}({x}_{0} | {x}_{t}, {f}).
\end{align}
With the loss, the network can focus more on learning how to generate text, rather than how to learn the image features.
During inference,
the probability likelihood is correspondingly computed as
\begin{align}
&\log p_{\theta}({x}_{0} | {x}_{t}, {y})'=\log p_{\theta}({x}_{0} | {x}_{t}, {f}) + \\& s(\log p_{\theta}({x}_{0} | {x}_{t}, {y}) - \log p_{\theta}({x}_{0} | {x}_{t}, {f})),
\end{align}
where $s$ denotes the guidance scale ($s = 1.17$ in experiments). If $s=1$,  the image-free process is effectively removed. If $s > 1$, the network will respect more of the image signals.

\section{Experiments~\label{sec:exp}}
\subsection{Setup}
\noindent\textbf{Dataset.} We conduct our experiments on the popular COCO dataset~\cite{Lin2014MicrosoftCC}. Specifically, the COCO dataset contains 123,287 images labeled with 5 captions for each, including 82,783 training images, 40,504 validation images and 40,775 images as test set for online evaluation as well. 
Following the widely-used “Karpathy” split~\cite{Karpathy2017DeepVA}, we use 113,287 images for training, 5000 images for validation, and 5000 for testing. 
For our diffusion model, our vocabulary size is 50,257, and the max length of each sentence is 20. During training, the weight decay is set to 0.01, and the learning rate first linearly warms up to 2e-4, and then drops to 0 following cosine decay.
The number of training epochs is 30, and the warm-up epoch is 5. The batch size is 512. Following the standard evaluation setup, we report the performances of our model and compare to other methods over five metrics: BLEU@4~\cite{Papineni2002BleuAM}, METEOR~\cite{Banerjee2005METEORAA}, ROUGE-L~\cite{Jain2018TwoCP}, CIDEr-D~\cite{Vedantam2015CIDErCI}, and SPICE~\cite{Anderson2016SPICESP}.

\vspace{2mm}
\noindent\textbf{Network backbone and pre-training.} ViT-B/16~\cite{Dosovitskiy2021AnII} from the pre-trained CLIP model~\cite{radford2021learning} is chosen as our image backbone, and the corresponding image patch size is 16. More specifically, ViT-B/16 has 12 transformer layers with 768 as the hidden size. 
For pre-training, following common practice~\cite{chen2020uniter,dou2021empirical},
we use a combined dataset including COCO~\cite{Lin2014MicrosoftCC}, Conceptual Captions (CC)~\cite{Sharma2018ConceptualCA}, SBU~\cite{Ordonez2011Im2TextDI}, and Visual Genome (VG)~\cite{Krishna2016VisualGC}. This results in roughly 4 million images with 10 million associated captions. We also pre-train the model based on ViT-L/16 with 15
epochs when comparing with state-of-the-art methods. For our diffusion model, the peak learning rate is 1e-4 with batch size 1024 during pre-training and 1e-5 with batch size 512 during fine-tuning. 
The diffusion model is randomly initialized and 
the learning rate of the well-initialized image encoder is reduced to 0.07 of the diffusion model.

\subsection{Analysis}
Due to the consideration of computation cost, we conduct ablation study based on the fixed image encoder and disable the image-free training unless explicitly specified. 
No pretraining is conducted, and all results are reported on the ``Karpathy'' validation set. 

\paragraph{Ablation on the key designs.}

As discussed in Section~\ref{sec:key_design}, we have four key components in our model. Table~\ref{tab:component} reports the ablation study with different combinations. 
Comparing row (c) and (a), we can see the Best-first Inference strategy can significantly boost the performance from 20.6 to 45.2 in CIDEr. Thus, it is crucial to keep the best predictions unchanged during the denoising process.
By appropriately masking the attentions with CAM, the performance can be improved from 45.2 (row c) to 97.5 (row e).
This suggests that it is necessary to prevent tokens from depending on \texttt{[MASK]} tokens. 
Otherwise, it may reduce the weights of other text tokens and lead to the reduction of the signal-to-noise ratio. 
The length prediction branch facilitates the model to find the end of the caption, 
and we can see the performance improved from 45.2 (row c) to 92.6 (row d).
Lastly, image-free training also helps improve the performance (116.7 to 117.8 from row f to row g).






\paragraph{Concentrated attention mask.}\label{sec:maskedattn}

In our proposed concentrated attention mask (CAM), we have two modifications: 
($i$) the text token does not depend on \texttt{[MASK]} tokens, denoted as T2M; 
($ii$) the \texttt{[MASK]} token does not depend on other \texttt{[MASK]} tokens, denoted as M2M. 
Table \ref{tab:maskatten} 
suggests that both T2M and M2M improve the performance and M2M provides much larger gain 
(from 92.6 to 115.6). 
The reason may be that \texttt{[MASK]} token's representation is the key 
to recover new tokens and it is beneficial not to focus on the other \texttt{[MASK]} tokens, which carry less information. 
We also experiment to enable CAM only at the first half or the second half denoising steps, as in the last two rows of Table \ref{tab:maskatten}.
In the first half ($t \ge 10$), more tokens are \texttt{[MASK]} and thus CAM impacts  more to the performance. 
The results also demonstrates CAM is an effective way to improve the performance.

\begin{table}
  \centering
  \scalebox{0.80}{
  \begin{tabular}{@{}l c c|c|c|c|c|c@{}}
    \toprule
    Method  & M2M & T2M & C & B@4 & M & R & S\\
    \midrule
    DDCap &&& 92.6 & 27.3  & 25.4 & 51.8 & 18.7  \\ 
    DDCap &&$\checkmark$& 94.1 & 27.2  & 25.7 & 52.1 & 18.9  \\ 
    DDCap  &$\checkmark$ && 115.6 & 34.2  & 27.9 & 57.2 & 21.4  \\
     DDCap  &$\checkmark$&$\checkmark$ & \textbf{116.7} & \textbf{34.6}  & \textbf{28.1} & \textbf{57.4} & \textbf{21.5}\\
     \midrule
    DDCap ($t < 10$)  &$\checkmark$&$\checkmark$ & 98.8 & 29.8  & 25.2 & 52.8 & 19.4\\
    DDCap ($t\geq10$)  &$\checkmark$&$\checkmark$ & 112.9 & 33.2  & 27.9 & 56.5 & 21.2\\
    \bottomrule
  \end{tabular}
  }
  \caption{
  Ablation study on our proposed concentrated attention mask (CAM). 
  ``M2M": the mask token does not depend on the other mask tokens. 
  ``T2M": the text token does not depend on mask tokens. 
  ``($t < 10$)": CAM is enabled at the second half of denoising steps. 
  ``($t\geq10$)": CAM is enabled at the first half of denoising steps where most of the tokens are mask tokens. 
  }  
  \label{tab:maskatten}
\end{table}

\begin{figure}[t]
   \centering
   \includegraphics[width=1\linewidth]{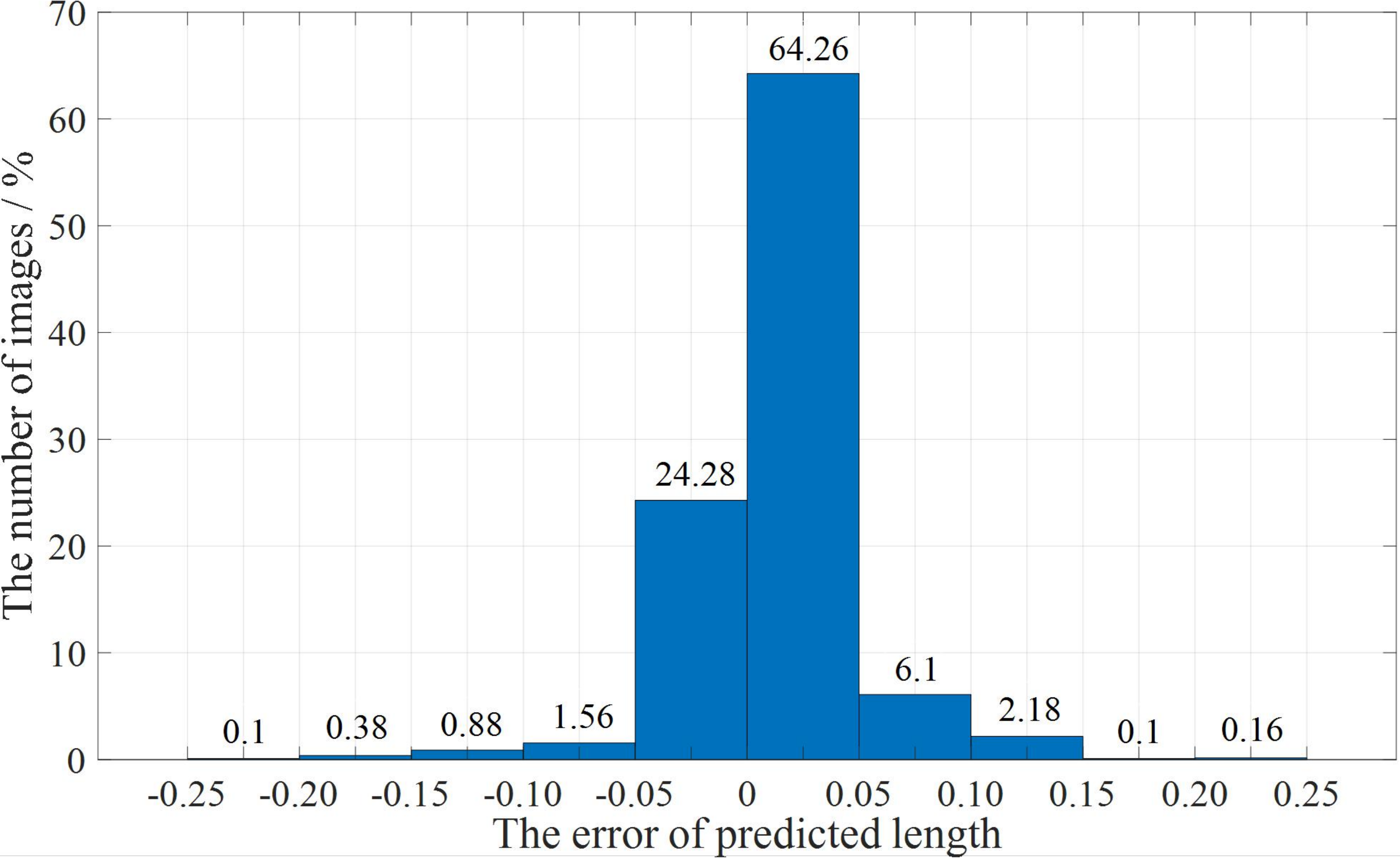}

   \caption{The accuracy of length prediction. }
   \label{fig:lengthpred}
\end{figure}

\paragraph{Error rate of the length prediction.}
Figure \ref{fig:lengthpred} shows the error distribution of the length prediction branch. 
The error is calculated as $(N_L-GT)/GT$, where $N_L$ denotes the predicted length and $GT$ is the ground-truth length. 
As we can see, most of the predictions are exactly correct, the error rate of 88.54\% images are less than 0.05. The max error rate is less than 0.25. This suggests the predicted length is reliable to determine the token length in decoding.

\paragraph{Embedding for time step $t$.}
We use the sinusoidal positional embedding
as in Eqn.~\ref{eqn:sin} to map the time step $t$
to an embedding representation. Another way is to use a learnable
embedding layer.
Table~\ref{tab:adaptivet} compares these two different methods. 
The results indicate the sinusoidal positional embedding
with appropriate rescaling factor can achieve better performance.

\begin{table}
  \centering
  \scalebox{0.88}{
  \begin{tabular}{@{}l|c|c|c|c|c@{}}
    \toprule
    Method & C & B@4 & M & R & S\\
    \midrule
    DDCap w/o Adaptive T & 115.5 & 34.2  & 28.0 & 57.3 & 21.3\\
    DDCap w/ Adaln T & 115.1  & 34.1 & 28.0 & 57.1 & 21.3\\
    DDCap w/ sin/cos T (4000) & 115.9 & 34.5  & 28.0 & 57.4 & 21.4\\
    DDCap w/ sin/cos T (6000) & 114.8  & 33.7  & 27.9 & 
57.0 & \textbf{21.5}\\
   DDCap w/ sin/cos T (8000) & \textbf{116.7}  & \textbf{34.6}  & \textbf{28.1} & \textbf{57.6} & \textbf{21.5}\\
   DDCap w/ sin/cos T (10000) & 115.1  & 34.0  & 27.9 & 
 56.9 & 21.3\\
    \bottomrule
  \end{tabular}
  }
  \caption{The influences of different ways of Adaptive LayerNorm layers. ``Adaln T" denotes the normal embedding layer, and `` sin/cos T" denotes the ``Sinusoidal PosEmb" layer.}
  \label{tab:adaptivet}
\end{table}





\begin{figure}[t]
	
	\begin{minipage}{1.0\linewidth}
		\centerline{\includegraphics[width=\textwidth]{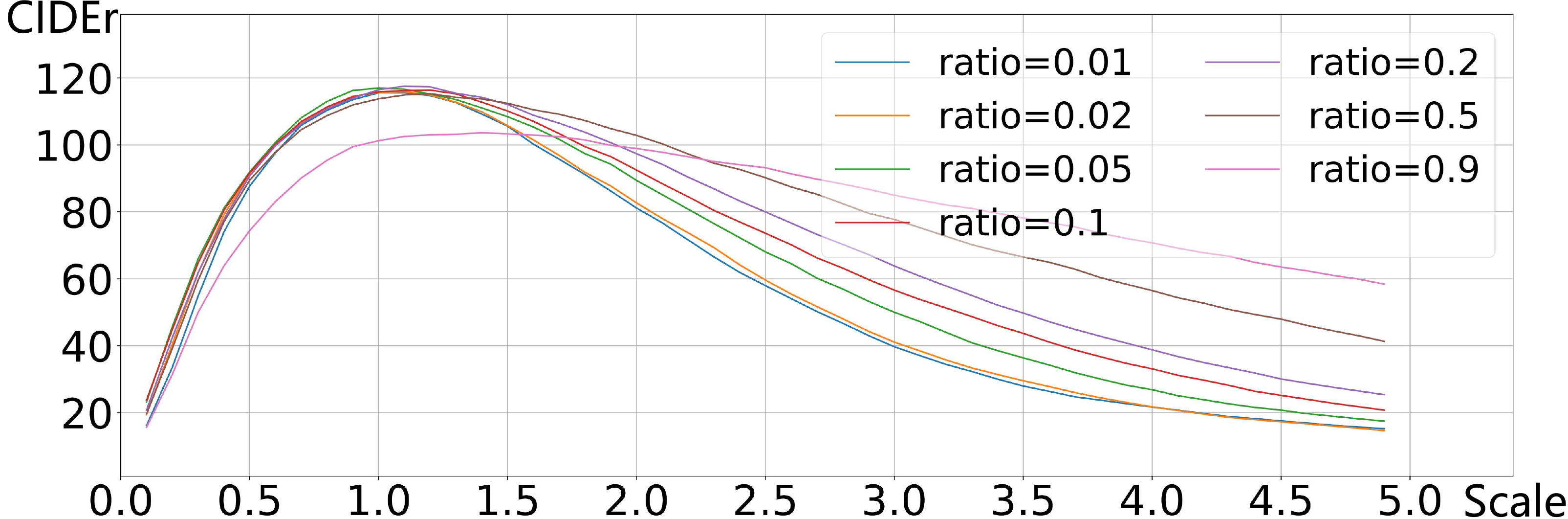}}
  \vspace{10pt}
	\end{minipage}

	\begin{minipage}{1.0\linewidth}
		\centerline{\includegraphics[width=\textwidth]{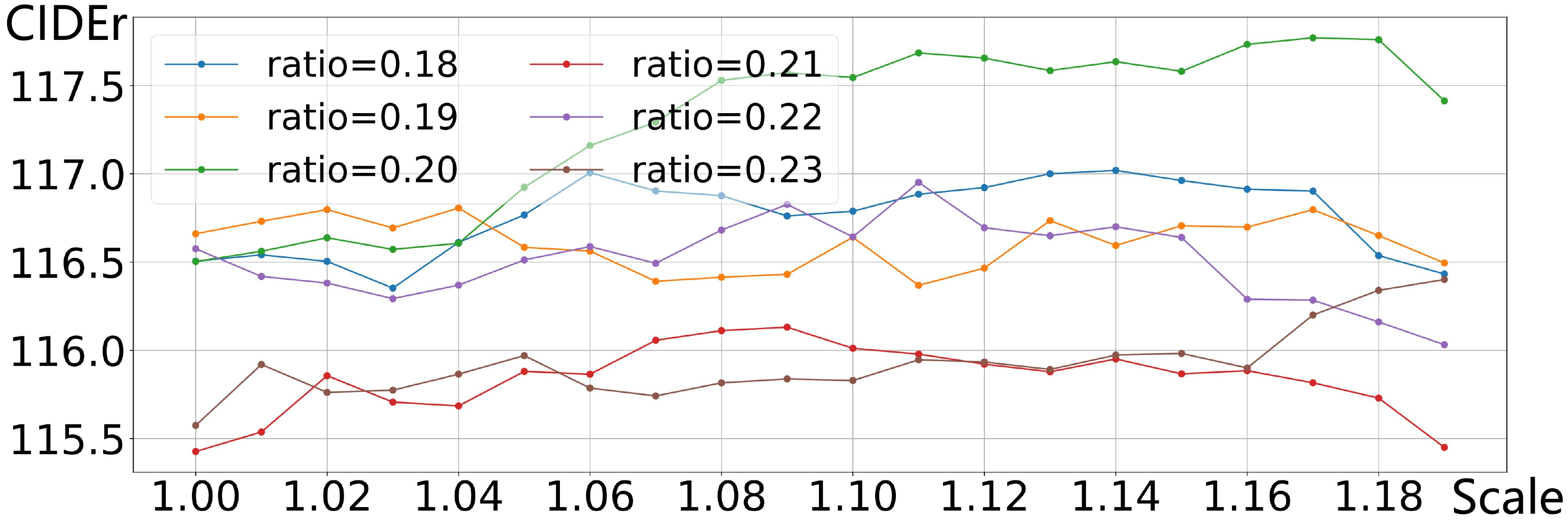}}
	 
	\end{minipage}
 
	\caption{The guidance scale $s$ and ratio $r$ of image-free training. Top: The interval of the scale is 0.1. Bottom: The interval of the scale is 0.01.}
	\label{fig:imagefree}
\end{figure}

\paragraph{Image-free training.}
The impact of hyper-parameters (\textit{i.e.}, $r$ and $s$) is shown in Figure \ref{fig:imagefree}. 
When the training ratio $r$ is 0.2 and the inference scale is 1.17, our model achieve the best performance. 
It is worth noting that the best guidance scale range is different to the one which is normally set to 5 for image generation~\cite{tang2022improved}

\begin{table}
  \centering
  \scalebox{0.9}{
  \begin{tabular}{@{}l|c|c|c|c|c@{}}
    \toprule
    Method  & C & B@4 & M & R & S\\
    \midrule
    AR Basline & 112.8  & 33.9 & 28.0 & 56.4 & 21.3\\
    Continuous Diffusion & 91.9 & 25.0 &25.0&51.1& 19.1 \\
    DDCap  & \textbf{117.8} & \textbf{35.0}  & \textbf{28.2} & \textbf{57.4} & \textbf{21.7}\\
    \bottomrule
  \end{tabular}
  }
  \caption{Comparison to an auto-regressive (AR) baseline and our implementation of a continuous diffusion model for captioning.}
  \label{tab:framecom}
\end{table}



\paragraph{vs AR approach and continuous diffusion model.}
We compare our proposed method with an auto-regressive (AR) baseline and our implementation of a continuous diffusion model in a controlled way. 
For AR approach, we use the same image encoder to extract the image representation, and re-use our diffusion transformer network as the decoder, such that the number of model parameters are similar. 
For the continuous diffusion model, we map the discrete text tokens to a continuous reprentation through the pretrained embedding layer of GPT-2, and use the diffusion transformer to recover these embeddings. 
The number of total step $T$ is set to 10,000, and the noise schedule is linear.
We find it helps with the following modifications: 
1) normalize the embedding layers with the mean and variance of all embedding vectors; 
2) during inference, each embedding vector $x_{t}$ is decoded to tokens and then re-embedded to vectors before estimating $x_{t-1}$.
The training epochs are set to 30, 100 and 30, for AR model, continuous diffusion model, and our discrete diffusion model, respectively. 

\begin{figure}[!t]
   \centering
   \includegraphics[width=1\linewidth]{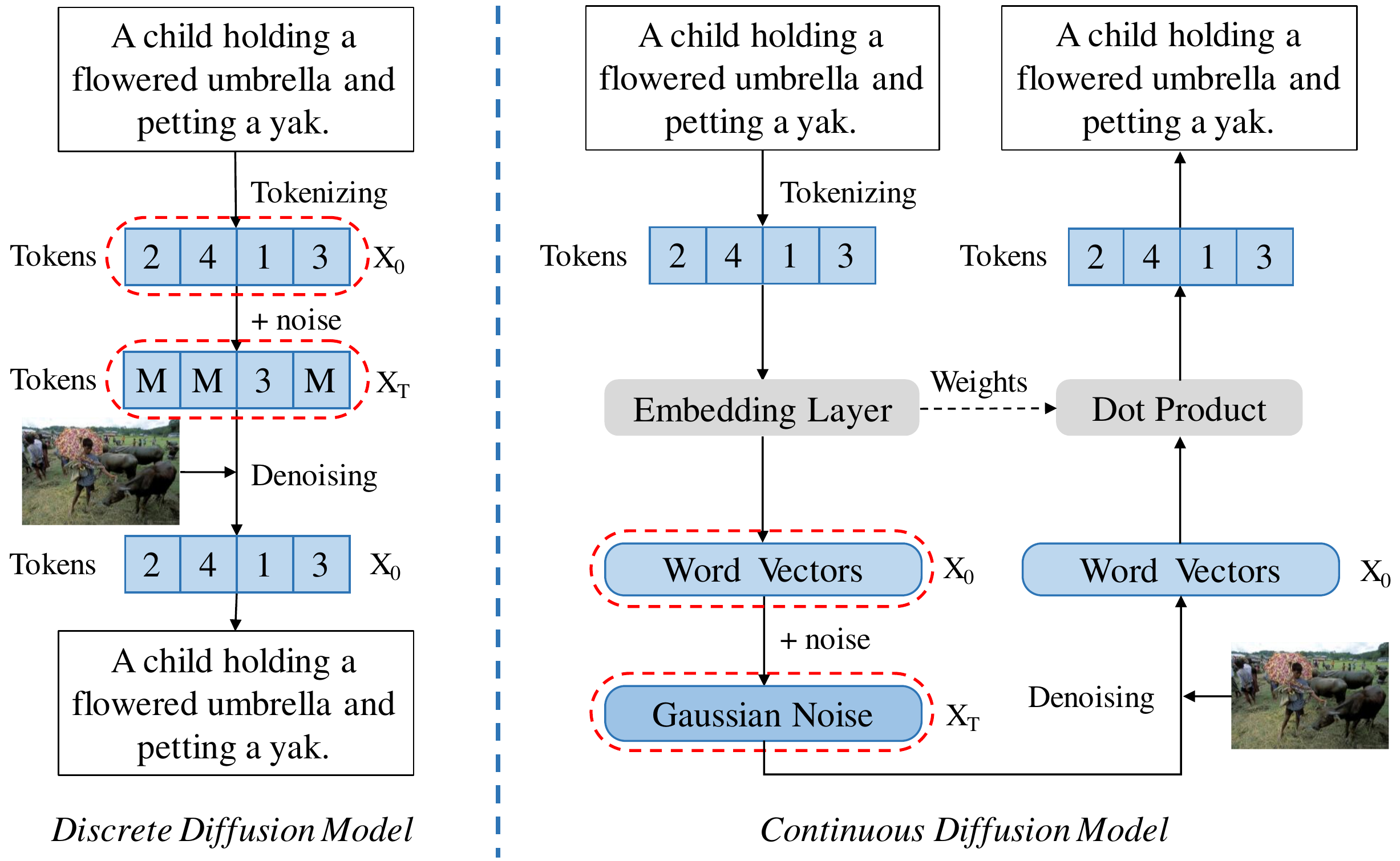}

   \caption{An overview of using \emph{discrete} and \emph{continuous} diffusion models for image captioning. Empirically, discrete diffusion achieves better performance, which is also the focus of this paper.}
   \label{fig:cont_vs_disc}
\end{figure}

The results are shown in Table~\ref{tab:framecom}, which indicates that our discrete diffusion model demonstrates better performance 
than auto-regressive model and the carefully designed continuous diffusion model. 
Compared with auto-regressive model, the discrete diffusion model can utilize bidirectional context in text. 
Compared with continuous diffusion model, discrete diffusion model aligns better as the text is naturally discrete. Moreover, the differences between continuous and discrete diffusion models are shown in Figure~\ref{fig:cont_vs_disc}.

\subsection{Comparison with Prior Arts}
Table~\ref{tab:compsota}  presents the comparison results on the COCO dataset. 
Our \modelname~shows competitive performance. Specifically, compared with non-autoregressive methods, our model achieves the best performance, which suggests the effectiveness of our key designs. Compared with auto-regressive methods, our performance is better than many methods and is comparable to ViTCap~\cite{fang2022injecting}. 
Hopefully, our study on the diffusion model can motivate more efforts on this direction.

\begin{table}[t!]
  \centering
  \setlength\tabcolsep{3pt}
  \scalebox{0.86}{
  \begin{tabular}{@{}l c c c c c c @{}}
    \toprule
    Method & \#Param. & \#Images & C & B@4 & M & S\\
    \midrule
    \multicolumn{7}{l}{\textbf{Auto-regressive models}}\\
    \midrule
    $\rm{UVLP}$~\cite{zhou2020unified}  & 111.7M & 4M & 116.9 & 36.5 & 28.4 & 21.2 \\
    $\rm{MiniVLM}$~\cite{wang2020minivlm}  & 34.5M & 14M & 119.8 &  35.6 & 28.6 & 21.6  \\
    $\rm{DistillVLM}$~\cite{fang2021compressing}  & 34.5M & 7M & 120.8 &  35.6 & 28.7 & 22.1 \\
    $\rm{UFO_B}$~\cite{wang2021ufo} & 0.1B & 4M & 122.8 & 36.0 & 28.9 & 22.2 \\
    $\rm{OSCAR_B}$~\cite{li2020oscar} & 0.1B+64M$^\dagger$ & 7M & 123.7 & 36.5 & 30.3 & 23.1\\ 
        $\rm{UNIMO_B}$~\cite{li2020unimo}  & - & 9M & 124.4 &  38.8 & - & - \\
    ViTCap~\cite{fang2022injecting} & 0.2B & 4M & 125.2 & 36.3 & 29.3 &  22.6 \\
    $\rm{VinVL_B}$~\cite{zhang2021vinvl} &  0.1B+0.2B$^\dagger$ & 6M & 129.3 & 38.2 & 30.3 & 23.6\\ 
    $\rm{GIT_B}$~\cite{wang2022git} & 129M & 4M & 131.4 & 40.4 & 30.0 & 23.0 \\
        \demph{$\rm{LEMON_B}$~\cite{hu2022scaling}} & \demph{111.7M} & \demph{0.2B} & \demph{133.3} & \demph{40.3} & \demph{30.2} & \demph{23.3}\\
    \demph{$\rm{SimVLM_B}$~\cite{wang2021simvlm}} & \demph{-} & \demph{1.8B} & \demph{134.8} & \demph{39.0} & \demph{32.9} & \demph{24.0}\\ 
    \midrule
    \multicolumn{7}{l}{\textbf{Non-autoregressive models}}\\
    \midrule
        $\rm{MNIC}$~\cite{gao2019masked}  & -  & - & 108.5 & 31.5 & 27.5 & 21.1\\
    $\rm{NAIC_{B,KD}}$~\cite{guo2020non}  & -  & - & 115.5 & 35.3 & 27.3 & 20.8\\
        $\rm{FNIC}$~\cite{fei2019fast}  & -  & - & 115.7 & 36.2 & 27.1 & 20.2\\
            $\rm{DDCap}$ (Ours)  & 280.1M  & 4M & \textbf{125.1} & \textbf{37.1} & \textbf{29.1} & \textbf{22.7}\\
    \bottomrule
  \end{tabular}
  }
  \caption{Performance comparison on COCO captioning Karpathy~\cite{Karpathy2017DeepVA} split with pretraining, where B@4, M, R, C denote BLEU@4, METEOR, ROUGE-L,
CIDEr and SPICE scores. CIDEr optimization is not used for all models. ($\dagger$) VinVL/OSCAR: the extra parameters are for
object detector. All of the results do not contain CIDEr optimization.}
  \label{tab:compsota}
\end{table}

\begin{figure}[t]
   \centering
   \includegraphics[width=1\linewidth]{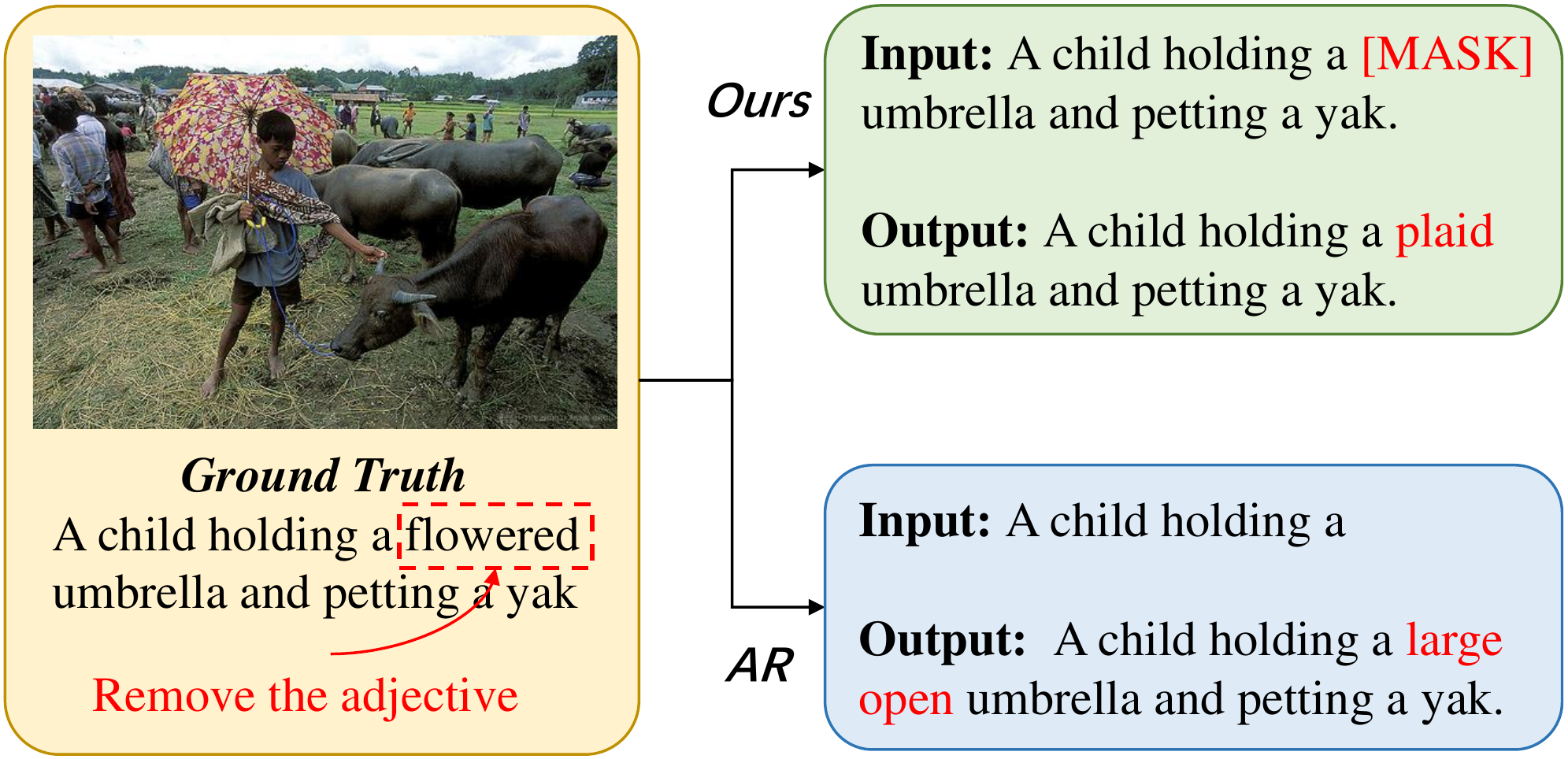}

   \caption{Visualization of the caption infilling task.}
   \label{fig:visualfill}
\end{figure}

\begin{figure*}[!t]
   \centering
   \includegraphics[width=1\linewidth]{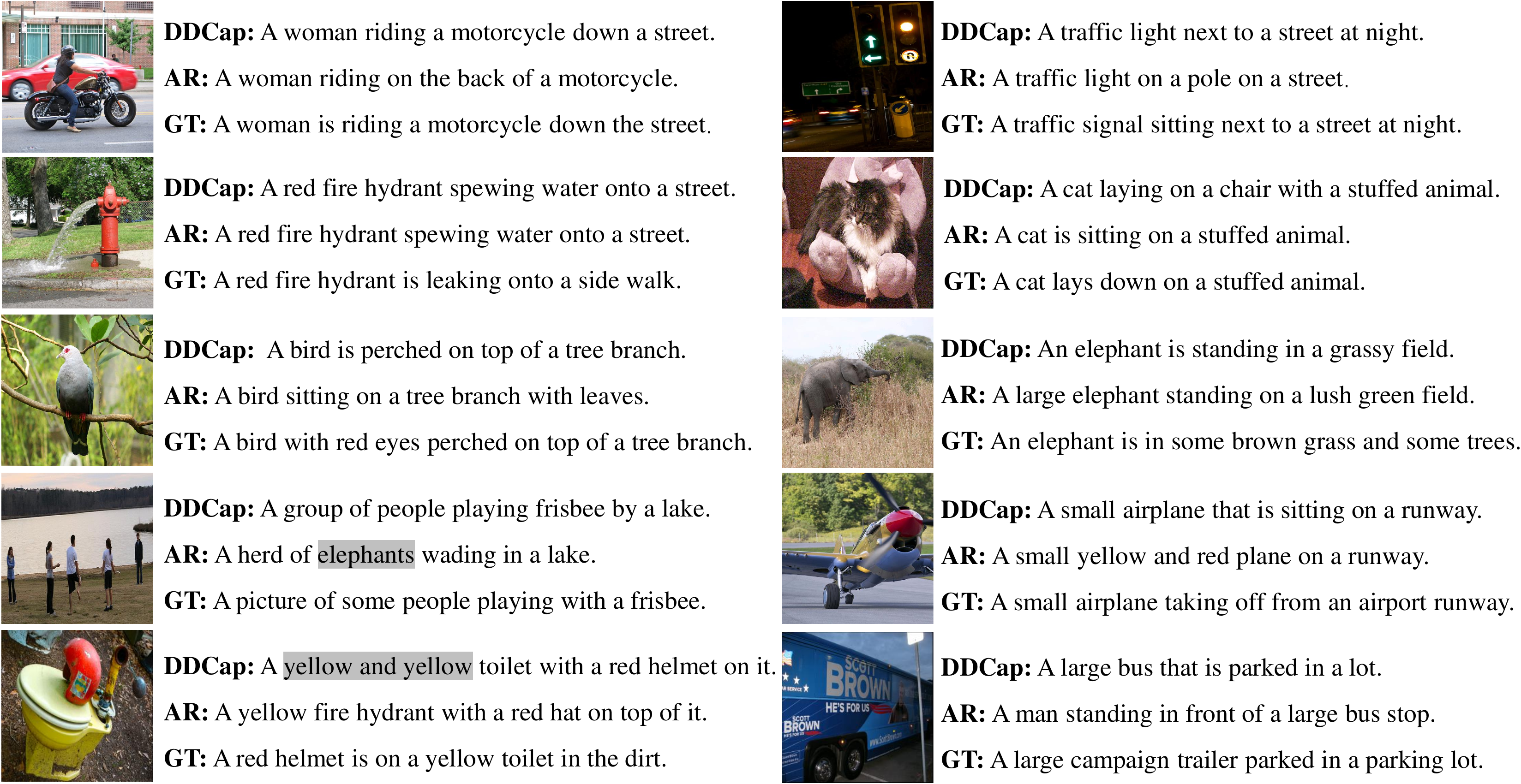}

   \caption{Visualization of our model and auto-regressive baseline on validation images of COCO dataset. The wrong parts of a caption are highlighted.}
   \label{fig:visualsota}
\end{figure*}

\begin{table}[!t]
  \centering
  \scalebox{0.87}{
  \begin{tabular}{@{}l|c|c|c|c|c|c@{}}
    \toprule
    Method & C & B@4 & M & R & S & CLIP-Score\\
    \midrule
    AR baseline & 203.5 & 76.3  & 49.3 &89.1& 36.5 & 75.7\\
    DDCap & \textbf{230.3} & \textbf{85.1}  &  \textbf{56.3} & \textbf{93.1} & \textbf{39.9} & \textbf{76.4}\\
    \bottomrule
  \end{tabular}
  }
  \caption{Comparison between our model and an auto-regressive baseline for the caption infilling task.}
  \label{tab:infill}
\end{table}

\begin{figure}[!t]
   \centering
   \includegraphics[width=1\linewidth]{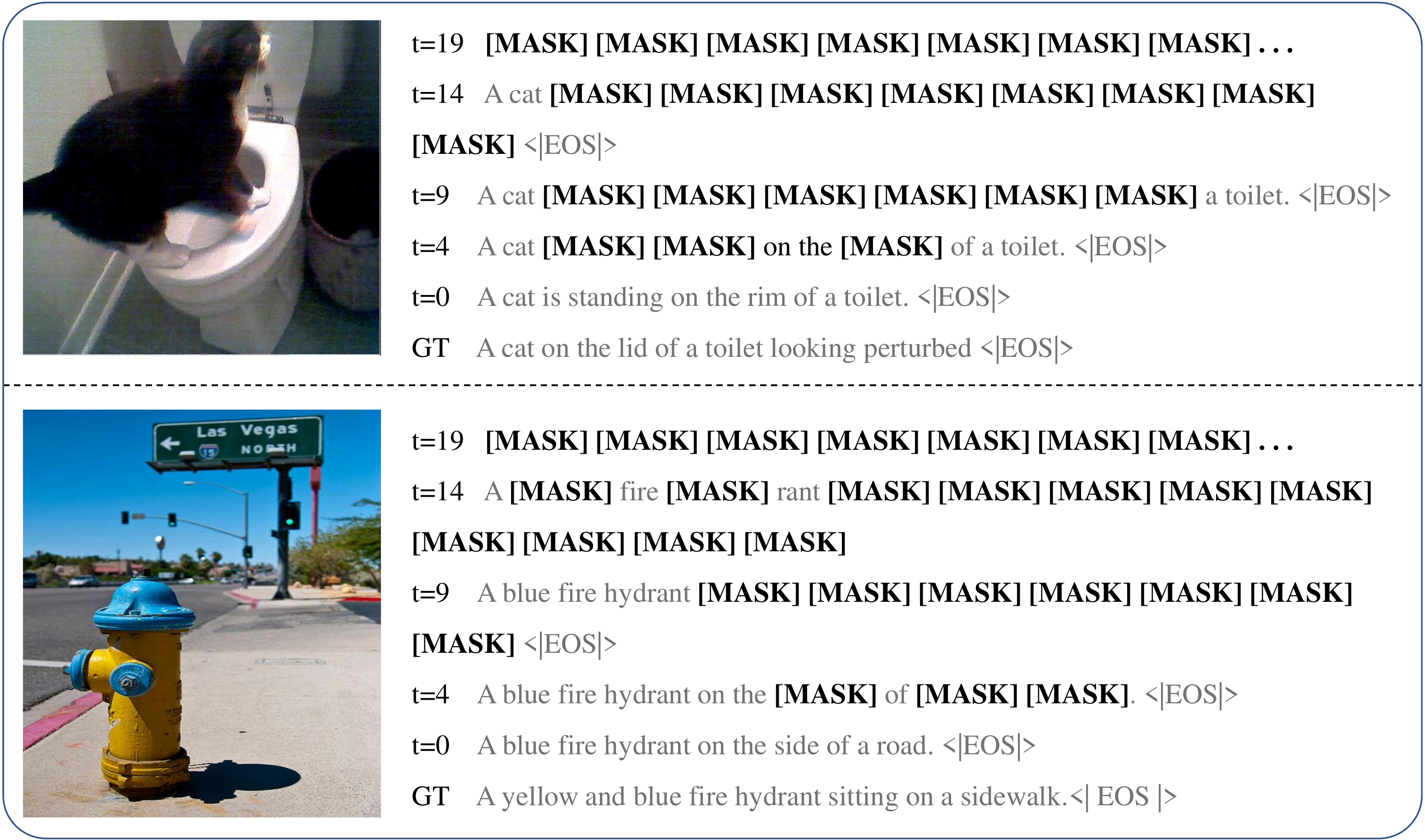}

   \caption{Visualization of the intermediate steps of the caption generation process with our model.}
   \label{fig:visualeachstep}
\end{figure}

\subsection{Caption Infilling Tasks}
As the text tokens are not decoded from a fixed left-to-right order, our approach is naturally suitable for the fill-in task task, which targets to fill in the empty words in a sentence as shown in Figure~\ref{fig:visualfill}.
To evaluate the performance, we remove all the adjective words on the “Karpathy” validation dataset.
The adjectives are detected by nltk\footnote{\url{https://www.nltk.org/}}.
The captions with no detected words are removed.
Then, the to-be-filled positions are filled with the mask token to start a diffusion process to recover the full sentence. 
As for evaluation metrics, we add the CLIP scores~\cite{hessel2021clipscore} for semantic matching. 
The comparison are shown in Table~\ref{tab:infill}, suggesting the bi-directional model can well solve the task.


\subsection{Qualitative Analysis}
Figure \ref{fig:visualeachstep} shows the text tokens in the intermediate steps. 
Empirically, simple objects and articles are predicted first, followed by preposition and adjectives. 
Finally, DDCap will add nouns to make the sentence grammatically correct. In other words, the way of describing the image roughly follows the order of first to describe the objects and then the relationship between the objects. 

Furthermore,  Figure \ref{fig:visualsota} shows prediction examples on the MSCOCO validation set, which shows reasonable prediction results. Compared with auto-regressive baseline, our DDCap has obvious advantages in recognizing objects. This advantages maybe come from our concentrated attention mask which provide more text context. 

However, our DDCap sometimes generates some repeated words. This issue may be handled by two ways: ($i$) we first get the caption, and then find the these repeated words. these repeated words will be re-masked and re-generated. ($ii$) the repeated words in a caption will be deleted. Then we get a new length of this caption without repeated words. According to the new length, we re-generate the whole caption.



\section{Conclusions~\label{sec:conc}}
This paper introduces a novel discrete diffusion model, termed DDCap, for image captioning. DDCap proposes several novel designs: length prediction, concentrated attention mask, and best-first inference. Ablation experiments under controlled settings demonstrate the effectiveness of each component. Results on COCO datast show that DDCap is comparable with state-of-the-art auto-regressive methods. In addition, we have introduced a new caption infilling task to highlight our advantages. For future work, we plan to perform larger-scale pre-training, and we hope our work can inspire more efforts on using diffusion models for text generation.
{\small
\bibliographystyle{ieee_fullname}
\bibliography{egbib}
}
\clearpage
\appendix
\twocolumn[\centering\section*{Appendices}]
\begin{table}[h]
  \centering
  \small
  \scalebox{1}{
    \begin{tabular}{l |c|c}
    \toprule
    \textbf{Hyperparameters} & \textbf{Image encoder} & \textbf{Diffusion model}\\
    \midrule
    Layers & 12 & 12\\
    Hidden size & 768 & 768\\
    Attention heads & 12 & 12\\
    Patch size & 16 * 16 & -\\
    Adaptive layer norm & - & sinusoidal \\
    Noise  schedule & - & linear schedule\\
    The total step $T$ & - & 20 \\
    \midrule
    Training steps & 30k & 30k \\
    Batch size & 512 & 512\\
    AdamW $\epsilon$ & 1e-8 & 1e-8\\
    AdamW $\beta$ & (0.9, 0.999) & (0.9, 0.999)\\
    Peak learning rate & 2e-5 & 2e-4\\
    Learning rate schedule & Cosine & Cosine\\
    Warmup steps & 6k & 6k \\
    \midrule
    Drop Path & 0.1 & -\\
    Weight decay & 0.01 & 0.01\\
    \midrule
    Image resolution & 224 & - \\
    Text length & - & 20 \\
    \bottomrule
    \end{tabular}  
  }
  \caption{
 Hyperparameters for training DDCap on COCO captioning without pretraining.
  }
  \label{tab:hyptrain}
\end{table}

\begin{table}[ht!]
  \centering
  \small
  \scalebox{1}{
    \begin{tabular}{l |c|c}
    \toprule
    \textbf{Hyperparameters} & \textbf{Image encoder} & \textbf{Diffusion model}\\
    \midrule
    Layers & 12 & 12\\
    Hidden size & 768 & 768\\
    Attention heads & 12 & 12\\
    Patch size & 16 * 16 & -\\
    Adaptive layer norm & - & sinusoidal \\
    Noise  schedule & - & linear schedule\\
    The total step $T$ & - & 20 \\
    \midrule
    Training steps & 142k & 142k \\
    Batch size & 1024 & 1024\\
    AdamW $\epsilon$ & 1e-8 & 1e-8\\
    AdamW $\beta$ & (0.9, 0.999) & (0.9, 0.999)\\
    Peak learning rate & 1e-5 & 1e-4\\
    Learning rate schedule & Cosine & Cosine\\
    Warmup steps & 47k & 47k \\
    \midrule
    Drop Path & 0.1 & -\\
    Weight decay & 0.01 & 0.01\\
    \midrule
    Image resolution & 224 & - \\
    Text length & - & 20 \\
    \bottomrule
    \end{tabular}  
  }
  \caption{
 Hyperparameters for pretraining DDCap.
  }
  \label{tab:hyppretrain}
\end{table}

\begin{table}
  \centering
  \scalebox{1}{
    \begin{tabular}{l |c|c}
    \toprule
    \textbf{Hyperparameters} & \textbf{Image encoder} & \textbf{Diffusion model}\\
    \midrule
    Training steps & 30k & 30k \\
    Batch size & 512 & 512\\
    Peak learning rate & 7e-6 & 1e-5\\
    Warmup steps & 6k & 6k \\
    \bottomrule
    \end{tabular}  
  }
  \caption{
 Hyperparameters for fine-tuning DDCap on COCO captioning.
  }
  \label{tab:hypfinetune}
\end{table}

\section{Ablation on the “Karpathy” test dataset}

In the main paper, we report the ablation study on 
“Karpathy”~\cite{Karpathy2017DeepVA} validation dataset. 
Here, we report the results on the test set in 
Table~\ref{tab:component_test}, in which we have 
consistent conclusions. 

\begin{table*}[t!]
  \small
  \centering
  \begin{tabular}{@{}c cccc|c|c|c|c|c@{}}
    \toprule   
    \#Row  & Best-first inference & CAM & Length Prediction & Image-free training & C & B@4 & M & R & S \\    
    \midrule
    a  &&&&& 20.4 & 7.1  & 18.8 & 34.3 &12.2  \\ 
    b  &&$\checkmark$&&& 39.0 & 10.5  & 20.3 & 38.9 & 14.1  \\ 
    c   &$\checkmark$ &&&& 45.4 & 20.4  & 26.9 & 47.2 & 21.4  \\
    d &$\checkmark$&&$\checkmark$& & 95.3 & 27.8  & 25.8 & 52.1 & 19.3\\
    e  &$\checkmark$&$\checkmark$&&& 97.2 & 27.7  & 28.0 & 53.8 & \textbf{21.8} \\
    f  &$\checkmark$&$\checkmark$&$\checkmark$&& 116.6 & \textbf{34.4}   & 28.0 & \textbf{57.2} & 21.3\\
    g &$\checkmark$&$\checkmark$&$\checkmark$&$\checkmark$& \textbf{117.9} & \textbf{34.4}  & \textbf{28.1} & 57.1 & 21.6\\
    \bottomrule
  \end{tabular}
  \caption{Ablation study on the effectiveness of each component on the “Karpathy” test dataset. CAM: concentrated attention mask.
  }
  \label{tab:component_test}
\end{table*}
\section{Hyperparameters}
The hyperparameters for training DDCap on COCO captioning are shown in Table~\ref{tab:hyptrain} with no pretraining.
The hyperparameters for pretraining DDCap are shown in Table~\ref{tab:hyppretrain}.

The hyperparameters for fine-tuning DDCap on COCO captioning are shown in Table~\ref{tab:hypfinetune}.
\end{document}